\documentclass[times]{simauth}
\usepackage{multirow}
\usepackage{tabularx}
\usepackage{moreverb}
\usepackage{graphicx}
\usepackage[driverfallback=dvips, colorlinks,bookmarksopen,bookmarksnumbered,citecolor=red,urlcolor=red]{hyperref}
\newcommand{\tabitem}{~~\llap{\textbullet}~~}
\usepackage{footmisc}
\newcommand\BibTeX{{\rmfamily B\kern-.05em \textsc{i\kern-.025em b}\kern-.08em
T\kern-.1667em\lower.7ex\hbox{E}\kern-.125emX}}

\usepackage{url}
\begin{document}

\runninghead{Choudhur Lakshminarayan and Tony Basil}

\title{Feature Extraction and Automated Classification of Heartbeats by Machine Learning}
\renewcommand{\thefootnote}{\fnsymbol{footnote}}

\author{Choudur Lakshminarayan\corrauth~and~Tony~Basil\footnote[2]{The work was undertaken when the author was at the Indian Institute of Technology (IIT) Hyderabad} }

\corraddr{Choudur Lakshminarayan, HP Enterprises, 14231 Tandem Boulevard, Austin, Texas, 78728.}

\begin{abstract}
We present algorithms for the detection of a class of heart arrhythmias with the goal of eventual adoption by practicing cardiologists. In  clinical practice, detection is based on a small number of meaningful features extracted from the heartbeat cycle.  However, techniques proposed in the literature use high dimensional vectors consisting  of morphological, and time based features for detection. Using  electrocardiogram (ECG) signals, we found smaller subsets of  features sufficient to detect arrhythmias with high accuracy. The features were found by an iterative step-wise feature selection method.  We depart from common literature in the following aspects: 1. As opposed to a high dimensional feature vectors, we use a  small set of features with meaningful clinical interpretation, 2. we eliminate the necessity of short-duration patient-specific ECG data to append to the global \emph{training} data for classification  3. We apply semi-parametric classification procedures (in an ensemble framework) for arrhythmia detection, and 4. our approach is based on a reduced sampling rate of $\sim$ 115 Hz as opposed to 360 Hz in standard literature.  
\end{abstract}

\keywords{ECG; Arrhythmia; feature selection; classification; sampling}

\maketitle

\section{Introduction}
{A}{ccording} to World Health Organization (WHO), cardiovascular diseases (CVDs) are the number one cause of death globally. An estimated 17.5 million people died from CVDs in 2012, representing 31\% of all global deaths \cite{WHO}. Figure \ref{fig:worldmap} shows the distribution of CVD related fatalities across the world. It is estimated that the number of fatalities due to CVDs will increase to 23.3 million by 2030.

\par With rapid urbanization and lifestyle changes, the prevalence of cardiovascular diseases is on the rise \cite{WHF}. This is putting additional burden on the healthcare industry due to insufficient infrastructure and rising cost of treatment. It is imperative that we find solutions that reduce number of fatalities, and ease the burden on healthcare providers. As the value of the emerging field of data science is ingrained in public consciousness and decision makers, it opens the opportunity for algorithms-supported decision making in the provision of healthcare.  
\par At present, cardiologists use an electrocardiogram (ECG) to monitor heart function of patients. Often detection of abnormal heart function is based on short-duration ECG patterns. It is well known that early and reliable detection of heart abnormalities require continuous and long term monitoring. To obtain heartbeat data over extended periods, the Holter monitor is used.  However, the Holter is invasive, cumbersome, and subject to interruptions in data collection. Lately, wireless body area networks (WBAN) \cite{sensornetwork} with non-invasive sensors that can record vital signs are being promoted to perform real time diagnosis \cite{alvarado}.  In recent years, there has been a proliferation of sensor based wearable devices such as Fitbit, Apple Watch, Garmin, and Samsung Smart watches.  These devices capture heart-rate data and other biological signals, but detection of anomalous patterns is still a developing technology. Also, startup companies such as LifeSync corporation (http://lifesynccorp.com) are developing wireless ECG data communication systems.  However,these are not proven  technologies in healthcare applications and require regulatory approval.  Technology adoption in healthcare requires fast and reliable detection. In the meanwhile, reliable methods for automated classification of bio-signals including heartbeats using ECG data is paramount to accelerate adoption in clinical practice. An automated classification and detection system must be accurate (few misclassification errors), stable (resistant to changes in the data), and use meaningful features.  Finally, a lower sampling rate is desirable in the context of WBAN as energy, data transmission and computational resource become critical. Therefore a minimal set of interpretable features, signals sampled at a lower rate, and a well trained algorithm with few underlying mathematical assumptions is desirable.  In this paper, we attempt to advance a solution that meets all the criteria mentioned above.  
\begin{figure}
  \begin{center}
    \includegraphics[width=85mm]{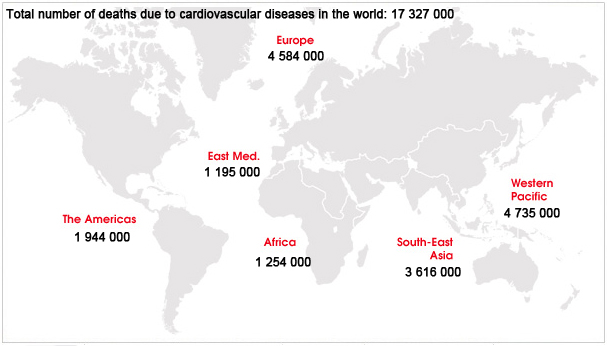}
  \end{center}
    \caption{Number of CDV related fatalities in each region across the world}
\end{figure}
\label{fig:worldmap}

\par Existing heartbeat classification algorithms generally focus on detecting Ventricular ectopic beats (VEBs). Consultations with cardiologists revealed that detection of VEBs may not be helpful in a clinical setting because they are not \emph{actionable}.  The cardiologist cannot provide a meaningful intervention in the presence of VEB patterns in ECG recordings.  However, detection of a different kind of arrhythmia, namely, Supraventricular ectopic beats (SVEBs) has significant clinical implications as early detection of SVEBs can prevent fatalities. 

\par In this paper, we propose techniques to detect two types of heartbeat arrhythmias - VEB (to align and compare with existing literature) and SVEB, with an emphasis on minimizing the burden on feature extraction and computational requirements of the algorithm. In existing heartbeat classification techniques, a large number of features are extracted from the ECG signal to train a classifier. While these techniques have been successful in accurate detection of VEB (considered unimportant by medical professionals), detection of SVEB may be improved.  But SVEB detection is prone to high false positive rates. In a real world setting, an algorithm with high rate of false positives is undesirable. The feature set is the cornerstone of statistical classification and therefore a judicious selection of a small set of meaningful features is important. Moreover, it is important to employ the most suitable classifier for detection purposes. We propose techniques to accurately classify heartbeats by \emph{training} a classifier using a small feature set (consisting of features in time domain, frequency domain and ECG morphology) obtained by employing a feature selection method \cite{Guyon} \cite{lannoy}.

\par The paper is organized as follows. Section II describes existing techniques for detecting various types of heart arrhythmias and briefly introduces our work. Section III describes the data, the \emph{training} and \emph{testing} sets, preprocessing, and best practices in handling the data. The various classifiers explored in this study are described briefly in Section IV. Section V summarizes 
features used in the literature and features we extracted and used for detection.  Section VI outlines the metrics used to assess algorithmic performance and Section VII describes the feature extraction and classification of VEB and SVEB followed by an analysis and discussion of experimental results. 

\section{Related Work} 

\begin{figure}
  \begin{center}
    \includegraphics[width=85mm]{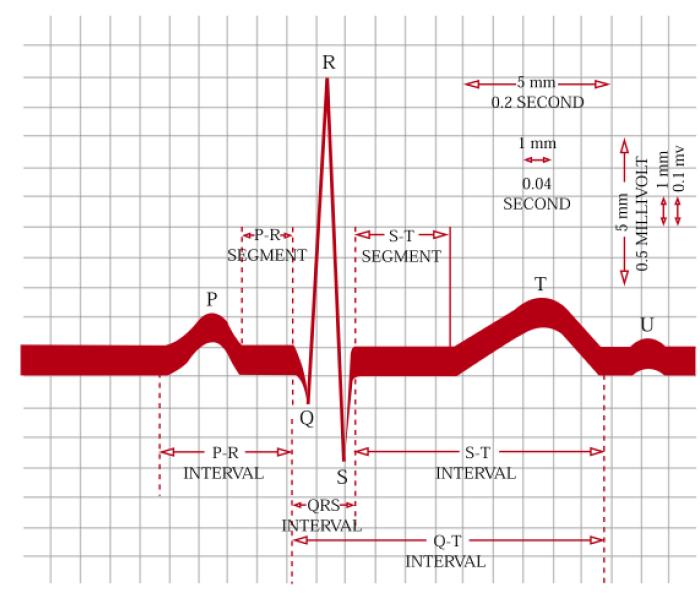}
  \end{center}
    \caption{Cardiac cycle of a typical heartbeat represented by the P-QRS-T wave form}
\label{fig:pqrst}
\end{figure}

Needless to say, classification of heartbeats is a challenging problem. This is due to near chaotic behaviors observed in heart abnormalities. ECG signals of heartbeats are characterized by features known as the P-wave, the QRS complex, and the T-wave (See Figure \ref{fig:pqrst}).  Typical features in classification include signal samples from the primitive features (P, QRS, T) and mathematical transformations thereof such as Fourier and Wavelets. A substantial number of these features are required for acceptable rates of detection. However, due to large variations within and between patients, the derived features are unstable. Therefore a careful selection of a small set of features related to heart function is essential. Several classifiers have been explored; chiefly the  Linear Discriminant Analysis (LDA), mixture of experts (MOE), Artificial Neural Networks (ANN), and Active Learning (AE).  

\par
\begin{figure}
  \begin{center}
    \includegraphics[width=85mm]{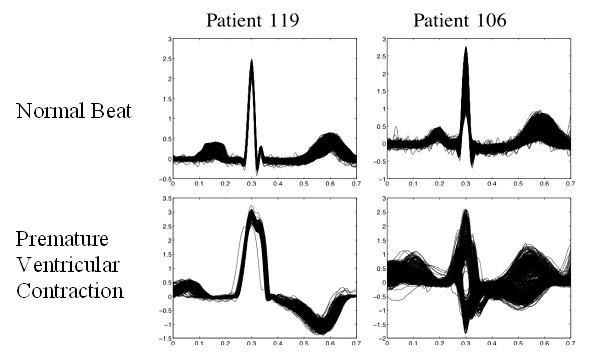}
  \end{center}
    \caption{Example of heartbeat shapes from the MIT-BIH data set. Each column represents a patient and each row the beats for that specific class. Note the variations in the beat morphology across patients as well as within a patient (Source Alvarado et al \cite{alvarado})}
\label{fig:heartbeats}

\end{figure}

 Heartbeat cycles are subject to statistical variation.  Figure \ref{fig:heartbeats}, shows variations in normal beats and variations in a type of arrhythmia known as premature ventricular contraction (PVC) between two patients.  Due to the erratic behavior of the signals, many of the time domain, frequency domain and ECG morphology features lack the consistency to represent heart function correctly. This affects learning in the training phase and the classifiers fail when applied to a signal from a new patient.  Hu et al \cite{Hu} used a “mixture of experts” model in which a global classifier and a local classifier are combined to make the classification decision. The local classifier is trained on patient specific labeled data and the global classifier uses the entire patient data set. A gating function is used to weight the classification decision from the global and local classifiers. Chazal et al (\cite{chazal1}, \cite{chazal2}) incorporate a similar local-global classifier mixture approach. However, Chazal et al \cite{chazal2} differs from Hu et al \cite{Hu} in feature extraction, and the number of patient specific heartbeats used for training the local classifier. In order to extract the various features of a heartbeat, it is crucial that we detect the boundaries of P wave, QRS and T wave. This is difficult because of signal noise.  Chazal et al \cite{chazal3} used a combination of feature sets from \cite{chazal1}, wherein the feature set for a particular heartbeat is selected based on the availability of information on the various wave boundaries (P wave, QRS and T wave).  Wiens et al \cite{wiens} proposed an active learning technique that reduces the number of patient-specific labeled data required to train a support vector machine classifier (SVM). Ince et al \cite{ince} proposed classification based on the ANN.  In brief, Ince et al \cite{ince} use wavelet coefficients and time-based features.  The wavelet coefficients are projected into a lower dimensional space via a linear transform-principal component analysis (PCA).  Needless to say, the resulting features lack clinical interpretation to be pertinent to the cardiologist.  The topology of the ANN is determined by varying the input and hidden layer nodes.  The ANN is trained using a small \emph{training} set and limited patient-specific data.  Alvarado et al \cite{alvarado} in a departure from traditional approaches used pulse based representations of signals using time based samplers such as Integrate and Fire (IF) model \cite{alvarado}. In \cite{BDA}, we compared the performance of LDA, QDA and artificial neural networks (ANN) in detecting Ventricular ectopic Beats (VEB). In \cite{ACS}, we focused on detecting Supraventricular ectopic Beats (SVEB) and proposed a classification technique based on the variations in the ECG morphology of SVEB’s. In \cite{EUSIPCO}, we propose new features and techniques to detect VEBs and SVEBs. In our approach, we prioritized SVEB over VEB as early detection of SVEB is of greater value to the cardiologist for prompt action than VEB [private communications with cardiologists]. Starting with a set of 31 time domain, frequency domain and morphological features (See Table \ref{table:table4}) and using the incremental wrapper approach (step-wise feature selection) \cite{Guyon} \cite{lannoy}, we determined a subset of four (using ANN) and eleven (using LDA) features that best capture heart function dynamics from a clinical standpoint.  Upon consulting with practicing cardiologists, we focused on a time domain feature known as pre-RR Interval (The time duration between the current heartbeat and the previous heartbeat). The pre-RR interval duration, while important in practice is subject to variations that renders it indistinguishable from normal beats.  To overcome, we apply a normalization technique to reduce its sensitivity to variation (See Section \ref{sec:EUSIPCO}). The pre-RR interval normalization yields a significant benefit in that it eliminates the necessity of a local classifier on patient-specific data as required by the local classifier in a “mixture of experts” framework. Lastly, the ANN classifier appears to be a suitable choice as it can model highly non-linear behaviors and it improved classifier performance. An important point to note is the resurgence of \emph{Deep Learning} as a mainstream technology in Big Data.  Therefore, the usefulness of ANN in heartbeat detection is a promising area for further examination.  

\section{Data Description} The labeled \emph{training} and \emph{testing} sets were created from the MIT-BIH \cite{physionet} arrhythmia database consisting of 48 real world patient recordings, each of 30 minute duration. A 22 patient subset consisting of 51,020 heartbeats was chosen as the \emph{training} set, while another subset of 22 patients consisting of 49,711 heartbeats was chosen as the \emph{testing} set. The remaining 4 patient recordings were not considered as they were on pacemakers and consist of only “paced,” (unknown type) heartbeats in compliance with American Association of Medical Instrumentation (AAMI) \cite{AAMI}.  The recordings were obtained by sampling the signals at the rate of 360 Hz using 11 bit precision A/D converter over a ±5mV range. We then down-sampled the signals to $\sim$ 115 Hz.  The lower sample rate was chosen to be on par with Alvarado et al \cite{alvarado}.  With \emph{training} set and \emph{testing} set clearly defined, a set of carefully chosen features were extracted from each heartbeat present in the two datasets.  We used the upper channel signal, which was acquired through a modified limb lead (MLL) with electrodes attached to the chest. The Database provides ECG signals with each heartbeat labeled by an expert cardiologist.  There are 20 types of beat annotations, as well as 22 types of non-beat annotations, including rhythm annotations.  The dataset consisting of 44 patients were divided into two sets DS1 and DS2 where DS1 was used for \emph{training} the algorithms and DS2 was used to evaluate the performance of various statistical classifiers. The sets, DS1, DS2 consist of the following patient records respectively.

DS1 = [101, 106, 108, 109, 112, 114, 115, 116, 118, 119, 122, 124, 201, 203, 205, 207, 208, 209, 215, 220, 223, 230]

DS2 = [100, 103, 105, 111, 113, 117, 121, 123, 200, 202, 210, 212, 213, 214, 219, 221, 222, 228, 231, 232, 233, 234]

Paced beats = [102, 104, 107, 217].

\subsection{ECG Filtering}
Acquisition of ECG recordings involves detecting very low amplitude voltage in a noisy environment. We preprocessed the recordings to reduce the baseline wander and the 60 Hz power line interference. To remove baseline wander, we passed the signal through median filters of window sizes 200ms and 600ms. It removes P-waves, QRS complexes and T-waves leaving behind the baseline wander. By subtracting the baseline wander from the original signal, a filtered signal is obtained. Furthermore, power line interference was removed by using a notch filter centered at 60Hz.

\subsection{Heartbeat Classes}
The MIT-BIH database has annotations for 20 different types of heartbeats in its repository. The annotation identifies the R-Peak for each heartbeat, where R-Peak represents the peak of the QRS complex.   In accordance with the AAMI standard \cite{AAMI}, we grouped the heartbeat types into 5 classes. They are Normal and bundle branch block beats (N), Supraventricular ectopic beats (SVEBs), Ventricular ectopic beats (VEBs), Fusion of normal and VEBs (F), and Unknown beats (Q). Table \ref{table:table1} shows the various heartbeat types listed in the MIT-BIH database. All the results presented are based on binary classifications (SVEB versus the rest, and VEB versus the rest) in compliance with AAMI \cite{AAMI} standard and consistent with studies reported in the literature, where the arrhythmia patterns (SVEB, VEB) are classified against the remaining heartbeat classes {N, S, F and Q}.

\section{Heartbeat Classification Methods}
Existing literature focuses on the detection of Ventricular ectopic beats (VEB) and Supraventricular ectopic beats (SVEB).  Although the proposed solutions can accurately detect VEB, they suffer from high false positive rates when applied to SVEB detection. So we lay importance on SVEB detection which is particularly relevant in clinical practice.  A variety of classifiers have been proposed in the literature.  Most notably, Linear Discriminant Analysis (LDA), Mixture of Experts (MOE), Artificial Neural Networks (ANN) and Active learning (AE).  We add to the existing set of classifiers by including Quadratic Discriminant Analysis (QDA), Artificial Neural Networks (ANN) ensembles.  We will touch on the highlights of each method briefly in the following.  

\begin{figure}
  \begin{center}
    \includegraphics[width=85mm]{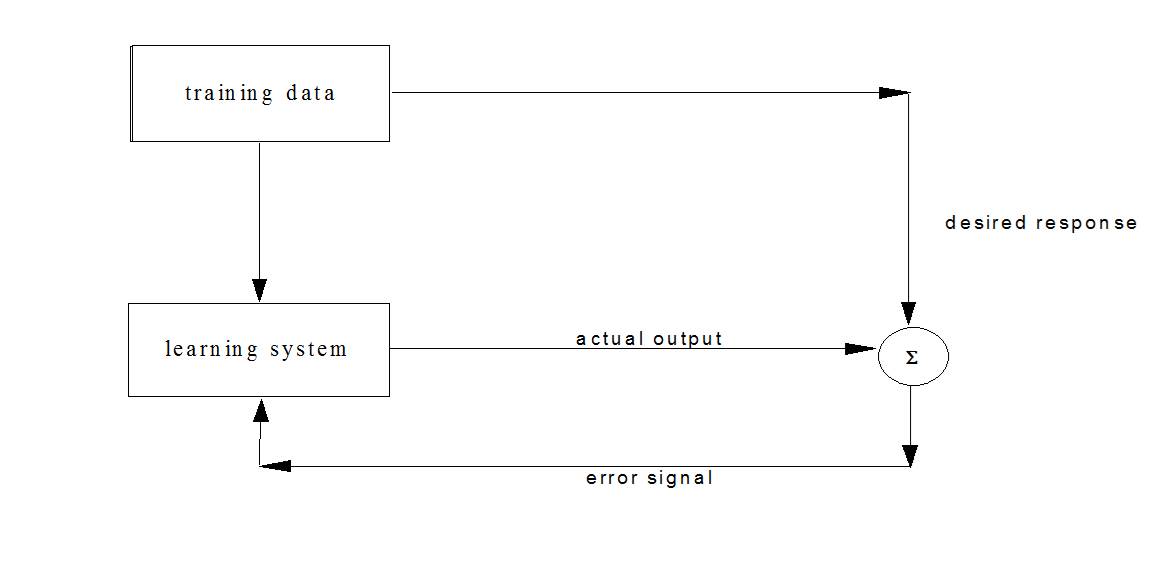}
  \end{center}
    \caption{Schematic illustration of an Artificial Neural Network}
\label{fig:neural}

\end{figure}

In the mainstream literature on arrhythmia detection, the LDA is the \emph{de facto} standard. Linear discriminant analysis assigns a \emph{p}-dimensional feature vector into one of \emph{k} classes based on a posterior probability of classification. LDA works best when the underlying probability density function of the data is truly Gaussian. By calculating the posterior probability of class membership of a new example from the \emph{testing} set, the example is classified into one of the\emph{ k} classes. Therefore, the classifier chooses the class with highest posterior probability \cite{chazal1}.  While the LDA assumes the class-conditional covariances are equal, QDA on the other hand differs from LDA in that the class-conditional covariances are treated unequal, therefore causing the decision function to be quadratic.  Artificial Neural networks (ANN) based on \emph{ Back propagation} algorithm is chosen often when it is difficult to mathematically express a relationship between the inputs (feature vector) and the outputs (classes). The attractiveness of ANN is that it does not require a precise mathematical parameterization of the relationship between the input features and the output classes. The ANN learns by iteratively minimizing an error function $(L_{2})$-sum of squares-and propagating the error back into the network (\emph{back propagation}) to update the weights until convergence is achieved.  For a pictorial description of backpropagation learning algorithm, see Figure \ref{fig:neural}. We use the ANN in an ensemble setting as another classifier.  In this approach, neural network models of various topologies are implemented.  The ensembles is an approach that utilizes an assembly of networks that work cooperatively to improve performance.  Outputs from the ensemble are averages computed from each member of the ensemble.  Typically, neural networks are specified by three configurable parameters.  They are; number of hidden layer nodes, the statistical distribution of the initial weights, and the learning rate $(\eta)$.  In practice, the hidden layer nodes are taken to be the sum of input layer and output layer nodes divided by 2 or 3.  The initial weights of the neural network are drawn from the uniform probability distribution with variance in the range of (0.1-0.5), and the learning rate ($\eta$) is in the interval (0.1-0.3). See \cite{elementstextbook} for an excellent treatment of ANN based models. In the next step, we run ANN models using different combinations of hidden layer nodes, learning rates, and the weights (uniform with variance between 0.1-0.3) and select the model with the best performance relative to the classifier metrics which are discussed  in the next section.  Upon selecting the best network topology, we rerun the ANN models iteratively 20 times where the initial weights are generated with varying seed values (the choice of twenty runs is based on experimental trials).  This step is to induce variability in the weights.  We average the metrics  averaged across the 20 runs.  We also report the standard error of the metrics to measure algorithm variability (stability).  The Mixture of experts (MOE) method employs a number of classifiers and outputs the result from the best performing classifier based on a decision rule.  In the heartbeat detection literature, the mixture involves a \emph{local} and a \emph{global} classifier, where the global classifier uses a \emph{training} set consisting of historical patient data and the local classifier uses patient-specific data  \cite{Hu}.  
Active learning is a semi-supervised learning paradigm in which the classifier interactively utilizes the \emph{training} data to obtain desired outcomes at new data points.  As labeled samples are fewer compared to unlabeled examples, classifiers can actively query the labeled data in the \emph{training} set for classification.   There is a need for identifying the best in class features/classifiers  since modifications to  classifiers proposed in literature \cite{ince, chazal2} did not achieve desired results. For instance, Chazal et al \cite{chazal1} employed a variant of LDA, where each class was assigned a weight to adjust for unequal sample sizes (the class of normal beats dominates the SVEB or VEB classes).  Therefore, in order to achieve desirable classification performance, it is imperative that we identify relevant features that are meaningful and a classifier that delivers.  So, our focus turned to identifying a small set of features and a dependable classifier.  We trained our classifiers to perform both five class classification and four class classification. In five class classification, each heartbeat is classified into one of the following classes -  Normal beats, SVEB, VEB, Fusion beats and Paced beats. In four class classification, the heartbeat is classified into one of the following classes - VEB, S1, S2 or Normal beats (See Section \ref{sec:ACS}). VEB represents Ventricular ectopic beats while the set of Normal beats is a union of normal beats, fusion beats and paced beats. On analyzing the morphology properties of SVEBs, two different morphological patterns emerged. The SVEB patterns were therefore divided into Classes S1 and S2 representing the two morphological patterns (See Figure \ref{fig:morphologypatterns}) and any heartbeat classified as either S1 or S2 is considered to be a SVEB. In accordance with AAMI recommendations \cite{AAMI}, we compute the performance measures for SVEB by analyzing the ability of a classifier to distinguish SVEB from non-SVEB. While computing the performance measures, the non-SVEB classes (Normal beat, VEB, Fusion beats and Paced beats) are treated as a single class. The performance measures are thus computed for binary classification rather than four(five) class classification. Same protocols were applied to compute performance measures for VEB. 

\section{Heartbeat Classification and Feature Sets}
As pointed out, classifiers in the literature are trained using a large number of features (eg. Wiens et al \cite{wiens} use a 67 dimensional feature vector!).  Chazal et al \cite{chazal1} created a set of eight feature sets FS1-FS8 (see \cite{chazal1} for details).  The features are a combination of RR-Interval (4), Heartbeat intervals (3), and Morphology (19) accruing to a total of 26 features (FS1, FS2, FS5, FS6).  The feature sets FS3, FS4, FS7, and FS8 contain 22 features each.  They analyzed the data using the eight feature sets.  Alvarado et al \cite{alvarado} used time domain features based on bin counts obtained using Integrate and Fire \cite{alvarado} algorithm.  While the results delivered using the features are impressive, their justification in practice do not stand to cardiological scrutiny. We constructed a set of 31 features  consisting of frequency domain,  time-domain (RR-interval, Heartbeat intervals), and morphology features [See Table \ref{table:table4}]. As the QRS complex is a dominant feature in the heartbeat cycle, we calculate the \emph{energy} ($\sum{x_{i}^{2}}$) within the QRS, QR, and RS segments.  These quantities give a measure of variability in the QRS complex.  Furthermore, we calculate the energy within the T-wave.  Armed with the 31 dimensional feature vector, we apply a dimensionality reduction method to identify a subset of features most correlated to the class label.  The feature selection method is known as Incremental wrapper approach \cite{Guyon}, \cite{lannoy}.  In statistical learning, wrapper methods treat feature selection as a search problem.  Various combinations of features are used to assess model performance and then select the best subset of relevant features, while discarding irrelevant ones.  Then the   subset is used to build a model for  upstream scoring.   The incremental wrapper scheme works similar to step-wise feature selection procedures common in generalized linear models and logistic regression in the statistics literature. The incremental wrapper method searches exhaustively for the best set of features in an iterative manner.  Given a \emph{k}-dimensional vector, theoretically the method involves building $2^{k}-1$ models.  For large \emph{k} the exhaustive search is infeasible as it turns out to be NP-hard.  So, a forward selection method is applied.  In this setting, the procedure begins with a classifier that delivers the best performance based on a single feature from the initial list of 31 features.  Subsequently, in the next steps, the feature whose addition to the current subset of feature(s) leads to the highest increase in prediction performance is retained. The procedure is iterated until further addition of features does not yield better performance.   In conjunction with ANN, we identified a set of 4 dominant features with good predictive capacity. Similarly, feature selection via LDA identified a 11 dimensional vector ( [See Table \ref{table:table5}]).  Results show that ANN trained using the 4 dimensional feature vector achieved significant improvement in detecting SVEB (See Table \ref{table:table6}).

\begin{figure}
  \begin{center}
    \includegraphics[width=85mm]{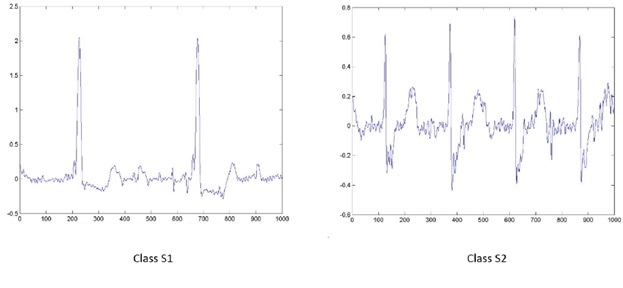}
  \end{center}
    \caption{Classes S1 and S2 representing two different morphological patterns of SVEB}
\end{figure}
\label{fig:morphologypatterns}

\section{Metrics}
A variety of metrics are used to assess classification performance.  Adhering to common practice in heartbeat classification,  results are reported in terms of sensitivity (Se), specificity (Sp), positive predictive value (PPV), false positive rate (FPR), and the F-Measure. The  F-measure is a harmonic mean of Se and PPV and by construction provides a conservative estimate of classifier performance.  The various measures are defined as follows: Se = $\frac{TP}{(TP+FN)}$, Sp =  $\frac{TN}{TN+FP}$, PPV = $\frac{TP}{(TP+FP)}$, FPR = $\frac{FP}{(TN+FP)}$ and F-Measure = $\frac{2*Se*PPV}{Se+PPV}$. where TP (True Positives): is the number of heartbeats that truly belong to class \emph{i} that are correctly assigned to class \emph{i}; FN (False Negatives) is the number of heartbeats belonging to class \emph{i} that are incorrectly classified to class \emph{j}, $({j \neq i})$; FP (False Positives) is the number of heartbeats belonging to class  \emph{j}, ($j \neq i$), that are incorrectly classified to class \emph{i}; and finally TN (True Negatives) is the number of heartbeats of class \emph{j } ($j \neq i$) that are correctly classified to class \emph{j}.    In passing, we note that sensitivity (Se) and (PPV) correspond to \emph{recall} and \emph{precision} respectively in the information retrieval literature.  
\section{Classification of arrhythmia patterns}
\subsection{Detection of SVEB based on ECG Morphology patterns}
\label{sec:ACS}
In existing literature, SVEBs are treated as a single class of heartbeat; with the assumption that SVEBs are morphologically similar. However, upon close examination, we observed that SVEB can be decomposed into two major categories based on the morphological patterns (See Figure \ref{fig:morphologypatterns}). So instead of representing SVEB as a single class, we subdivided SVEB into two classes S1 and S2, each representing a pattern.  Chazal et al \cite{chazal1} used datasets FS3 (22 features) and FS1-FS5 (26 features) which predominantly consist of ECG morphology features (see Table IV in \cite{chazal1}).  We were able to reconstruct FS1, FS3, and FS5 to make apples to apples comparison of the results from the two approaches.  The classification methodology involved four classes; S1 and S2 (derived from the single SVEB class), VEB (V) arrhythmitic rhythms,  and a fourth class (N) which is a union of  Normal, Fusion and Paced beats. The features in FS3 are:  RR interval and ECG morphology features respectively. The RR interval features are pre-RR interval, post-RR interval, average RR interval and local average RR interval. Pre-RR interval is the time interval between the peak amplitude (R-Peak) of previous heartbeat and the peak amplitude of current heartbeat. Post-RR interval is the time interval between the peak amplitude of current heartbeat and the peak amplitude of next heartbeat. ECG Morphology features include fixed interval morphology features from the QRS complex and T wave. In total, we extracted 10 features from the QRS complex and 8 features from the T wave respectively. These features were combined with RR interval features to form a 22 dimensional feature vector from each heartbeat in the 30 minute recording of all the patients. Similarly we extracted a 26 dimensional feature vector from the sets (FS1,FS5).  Furthermore we included the first three minutes of patient 232 to augment the representation for Class S2 in the \emph{training} set.  The results demonstrate that division of SVEB into S1 and S2 shows improvement in the metrics Se, FPR, SP, and the F-measure, while PPV is slightly lower (See Table \ref{table:table3} for details of the results).  The results from using the dataset FS3 (22 features) is shown in Table \ref{table:table2}. Table \ref{table:table2} reports the performance achieved by the four class (S1, S2 V, N) LDA classifier in detecting SVEB using the \emph{training} and \emph{testing} sets DS1 and DS2. We achieved a sensitivity of 78.8\% and PPV of 42\% which is comparable to the sensitivity and PPV reported by Chazal et al \cite{chazal1} (75.9\% and 38.5\% respectively). However, it underperforms compared to Chazal et al \cite{chazal2}, Alvarado  et al \cite{alvarado} and Wiens et al \cite{wiens} (See Table \ref{table:table2}).  Clearly, there is an improvement in the sensitivity measure. Our hypothesis is that it is perhaps due to the absence of  T-wave duration which is a significant predictor of SVEB.  In spite of the marginally improved results, it is our view that the decomposition of SVEB into the classes S1 and S2 needs critical examination and extracting features unique to the two types of patterns will further improve the results.  In the next section, we consider feature selection and classification to improve performance.

\subsection{Detection of SVEB using feature selection methods}
\label{sec:EUSIPCO}
To improve on the performance presented in the previous section, we approached the problem along two directions; feature selection and reduced sampling rate.  The lower sampling rate is motivated by the results by Alvarado  et al \cite{alvarado} who sample the signal at 117 Hz as opposed to the standard 360 Hz.  We first sample the signal at $\sim$ 115 Hz and invoke a feature selection process called the  the incremental wrapper approach.  The incremental wrapper approach is similar to step-wise regression techniques \cite{stepwise} which involves feature selection incrementally to determine the best set of features that are highly correlated with the output variable (deliver best performance). At the lower sampling rate of 115 Hz, we extract 4 samples from the QRS complex (two samples from either side of the R-peak as opposed to 10 features from QRS duration by \cite{chazal2}), features from the frequency domain, and time-duration features totaling 31 (See Table \ref{table:table4}).  Furthermore, in a preprocessing step,  we transformed the pre-RR interval duration by a normalization method to distinguish normal function from an SVEB arrhythmia.  The Normalized pre-RR Interval was computed by dividing the pre-RR interval of a heartbeat using the average pre-RR interval of normal beats (Any heartbeat other than SVEB, VEB, fusion and paced) in the neighborhood of that heartbeat. Since the Normal beats surrounding a heartbeat is not known a priori, the technique involves the detection of normal beats before computing the average pre-RR interval.  This was a significant finding as it improved performance considerably. Using LDA in conjunction with Incremental wrapper approach, a 11 dimensional feature vector was identified (See Subset 1 in Table \ref{table:table5}). Similarly, using ANN, the incremental wrapper algorithm produced a 4 dimensional feature vector (See Subset 2 in Table \ref{table:table5}). Practicing Cardiologists confirmed that these are indeed the first order features they use to identify SVEB patterns.  Among all the classifiers tested, an ensemble of ANN models achieved the best performance. An ensemble of 20 ANN models were used. The 20-run average is reported because we did not see significant change in the metrics beyond the 20 runs.  The average Sensitivity, PPV, and F-measure was calculated across the 20 runs to assess performance.  Finally, the classification results from the 5-class ANN was consolidated into 2 classes. The results are summarized in Table \ref{table:table6}. Due to space constraints, the results for specificity (Sp) is not reported. However, statistics related to specificity (Sp) is available at a link provided at the end of Section \ref{sec:Results}.

\subsection{Results}
\label{sec:Results}
The classification performance of the various algorithms is summarized in Table \ref{table:table6}. First five rows is a compilation of results from the literature, while the $\textbf{ last three rows}$ (in bold) represent results obtained using LDA, QDA, and ANN based on our proposed modifications. Column 1 identifies prior techniques and our proposed method(s), Columns 2, 3 and 4 represent sensitivity (Se), positive predictive value (PPV) and F-Measure respectively when the classification involves SVEB; while columns 5, 6, and 7 represent sensitivity (Se), positive predictive value (PPV), and F-Measure respectively when classification involves VEB. We especially call attention to  classification of SVEB type arrhythmia. 
Excluding Wiens et al \cite{wiens} (who use  a 67 dimensional feature vector), ANN in an ensemble setting performs the best relative to the F-measure, PPV.  The sensitivity (Se) while lower it is within range of the other methods.  But noticeably, PPV is startlingly improved compared to the competitors.  We believe the lower sampling rate eliminated some of the variation among the four combined  (non-SVEB) classes, and so false positives rate was reduced.  Furthermore, the sensitivity of 87.17 is close to 86.19 achieved by Alvarado et al \cite{alvarado} who also sample at $\sim$ 115 Hz.  We also report the standard errors (s.e) of the metrics (within parentheses).  We can see that clearly the algorithm is stable as the variance is a tiny fraction of the average metric values.  It confirms the reliability of the results reported.  Interestingly, LDA does well relative to the three metrics, followed by the Mixture of Experts (MOE).  The mixture is a combination of LDA and QDA.  The other classifiers (ANN, and ANN Ensembles) were not added to the mixture due to the incompatibility of classification criteria; i.e., LDA/QDA use posterior probabilities, while the others use the sum of squares of error.  We are exploring enriching the mixture by all the classifiers by suitable criterion common across all classifiers.  The results for the VEB type of arrhythmia self explanatory.  
It is noted that detection methods in a real-world setting are not intended to replace the cardiologist, but to serve as aid and reduce analytical cycle-times. The main failure of most machine learning techniques in the real-world are false positive rates, arcane feature sets, and esoteric machine learning algorithms.  Excessive false alarm rate cause practitioners to disregard real threats.  A procedure that delivers high sensitivity and low false positive rate is highly desirable and that is what we accomplished to deliver in this paper.  Finally, We encourage the reader to visit https://sourceforge.net/projects/ecganalysis/ \cite{sourceforge} for details, experimental results, MATLAB code, and references.

\section{Conclusion}
In conclusion, a judicious choice of the features meaningful to the cardiologist shows a measurable impact on the detection of some common types of heart arrhythmia (SVEB, VEB). The usage of the incremental wrapper approach helped to identify important features that are related to heart function while controlling for the dimensionality of the feature vector.  Our proposed solution eliminated the requirement of patient-specific labeled data. The application of the ANN classifier appears to have captured the non-linear behavior inherent in heart function. The down-sampling of the signal is important especially in the context of  big data and Internet of Things (IoT) that includes signal streams from a large population of patients which may lead to bandwidth constraints.  It is envisioned that these algorithms can be used in clinical settings as an assistive aid to cardiologists to accelerate the tedious process of examination and analyses of electronic cardiograms (ECG) charts.  As next steps, we are exploring enhancements to the Mixture of Experts approach that utilizes different sets of features for each type of arrhythmia and consists of a competitive network of different types of algorithms (experts) each with a different classification criterion to enhance detection and classification.  Furthermore, consultations with cardiologists revealed the significance of P wave in the heart cycle. However, due to the difficulty in detecting P waves with current methods, the feature did not yield significant improvement in performance.  P-wave feature extraction and  detection is therefore the critical step.  We are considering multiple approaches to accurate P-wave feature extraction.  Modeling the  dichotomous SVEB (S1 and S2) requires additional exploration and research to improve detection rates.

\begin{table*}[]
\centering
\caption{MIT - BIH heartbeat group and the corresponding AAMI Standard heartbeat class (Source Alvarado et al \cite{alvarado})}
\label{table:table1}
\def\arraystretch{1.5}%

\begin{tabular}{lll}
\hline
\hline
\bf MIT BIH heartbeat group                                                                                                                                                     & \bf AAMI heartbeat class            & \bf Beats \\ \hline
\begin{tabular}[c]{@{}l@{}}Normal beat\\ Left bundle branch block beat\\ Right bundle branch block beat\\ Atrial escape beats\\ Nodal (junctional) escape beat\end{tabular} & N: Normal beat                   & 90631 \\ \hline
\begin{tabular}[c]{@{}l@{}}Atrial premature beat\\ Aberrated atrial premature beat\\ Nodal (junctional) premature beat\\ Supraventricular premature beat\end{tabular}       & S: Supraventricular ectopic beat & 2781  \\ \hline
\begin{tabular}[c]{@{}l@{}}Premature ventricular contraction\\ Ventricular escape beat\end{tabular}                                                                         & V: Ventricular ectopic beat      & 7236  \\ \hline
Fusion of ventricular and normal beat                                                                                                                                       & F: Fusion beat                    & 803   \\ \hline
\begin{tabular}[c]{@{}l@{}}Paced beat\\ Fusion of paced and normal beat\\ Unclassified beat\end{tabular}                                                                     & Q: Unknown beat                  & 8043  \\ \hline
\hline
\end{tabular}
\end{table*}

\begin{table}[]
\centering
\caption{Classification performance of LDA}
\label{table:table2}
\def\arraystretch{1.5}%

\begin{tabular}{llll}
\hline
\hline
\multirow{2}{*}{{\bf Methods}} & \multicolumn{3}{c}{{\bf SVEB}}       \\

                               & {\bf Se} & {\bf PPV} & {\bf F-Measure} \\
                               \hline
Chazal et al \cite{chazal1}             & 75.9     & 38.5      & 51.08         \\
Chazal et al \cite{chazal2}            & 87.7     & 47        & 61.20         \\
Alvarado et al \cite{alvarado}         & 86.19    & 56.68     & 68.38         \\
Wiens et al \cite{wiens}              & 92       & 99.5      & 95.60         \\
{\bf Proposed LDA}                 & 78.8     & 42        & 54.79        \\
\hline
\hline
\end{tabular}
\end{table}

\begin{table*}[]
\centering
\caption{Comparison with state of the art classification techniques}
\label{table:table3}
\def\arraystretch{1.5}%
\begin{tabular}{lllllp{2cm}llllll}
\hline
\hline

\multirow{2}{*}{{\bf Methods}}                     & \multicolumn{5}{c}{{\bf SVEB}}       & \multicolumn{5}{c}{{\bf VEB}}        \\
                                                   & {\bf Se} & {\bf PPV} & {\bf FPR} & {\bf SP} & {\bf F-Measure} & {\bf Se} & {\bf PPV} & {\bf FPR} & {\bf SP} & {\bf F-Measure} \\
\hline
Chazal et al \cite{chazal1}                                 & 75.9     & 38.5     & 4.7 & 95.3 & 51.08        & 77.7     & 81.9  & 1.2   & 98.8 & 79.74          \\
Chazal et al \cite{chazal2}                                & 87.7     & 47    & 3.8 &  96.2  & 61.20         & 94.3     & 96.2  & 0.3 & 99.7   & 95.24         \\
Alvarado et al \cite{alvarado}                              & 86.19    & 56.68  & 2.55 & 97.45  & 68.38         & 92.43    & 94.82   & 0.4 & 99.6  & 93.60          \\
Wiens et al \cite{wiens}                                  & 92       & 99.5   & 0.0 & 100.0   & 95.60         & 99.6     & 99.3   & 0.1 & 99.9   & 99.44         \\

\bf{Proposed LDA}                              			& 89.1     & 36.8    & 4.1 & 95.9    & 52.09         & 82.04     & 78.25   & 1.59 & 98.41  & 80.1         \\
      \hline
            \hline

\end{tabular}
\end{table*}

\begin{table*}[]
\centering
\caption{List of features extracted from the ECG signal}
\label{table:table4}
\def\arraystretch{1.5}%
\begin{tabular}{llll}
\hline
\hline
\multicolumn{4}{l}{{\bf Features}}                                                                  \\
\hline
 \tabitem & Pre-RR Interval                       &    \tabitem  & T wave duration                           \\
 \tabitem & Post RR-Interval                       &    \tabitem  & Energy of QRS complex                     \\
 \tabitem & Average RR-Interval                    &   \tabitem  & Energy of QR segment                      \\
 \tabitem & Local Average RR-Interval                &  \tabitem  & Energy of RS segment                      \\
\tabitem  & QRS duration                              & \tabitem  & Energy of T wave                          \\
 \tabitem & QR duration                                & \tabitem  & Maximum Fourier coefficient of QR segment \\
 \tabitem & ECG Morphology of QRS complex (4 features) & \tabitem  & Maximum Fourier coefficient of RS segment \\
 \tabitem & ECG Morphology of T wave (9 features)     & \tabitem  & Maximum Fourier Coefficient of QRS complex \\
 \tabitem & P wave flag                              &  \tabitem  & Amplitude of R Peak                       \\
\tabitem  & Normalized pre-RR Interval               &  \tabitem  & RS duration                            \\
  \hline
  \hline
\end{tabular}
\end{table*}

\begin{table*}[]
\centering
\caption{List of features selected using incremental wrapper approach}
\label{table:table5}
\def\arraystretch{1.5}%
\begin{tabular}{llll}
\hline
\hline

\multicolumn{2}{l}{{\bf Subset 1 (11 dimensions)}} & \multicolumn{2}{l}{{\bf Subset 2 (4 dimensions)}} \\
\hline

   \tabitem & Normalized pre-RR Interval                   &    \tabitem & T wave duration                             \\
   \tabitem & Post RR-Interval                             &    \tabitem & Amplitude of R Peak                         \\
   \tabitem & T wave duration                              &    \tabitem & Maximum Fourer Coefficient of QRS complex   \\
   \tabitem & Energy of T wave                             &    \tabitem & Normalized pre-RR Interval                  \\
   \tabitem & ECG Morphology of QRS complex (4 features)   &    \tabitem &                                             \\
   \tabitem & Maximum Fourier coefficient of RS segment    &    \tabitem &                                             \\
   \tabitem & QRS duration                                 &    \tabitem &       \\
   \tabitem & Amplitude of R Peak                                 &    \tabitem &       \\

   \hline
         \hline                            
\end{tabular}
\end{table*}

\begin{table*}[]
\centering
\caption{Comparison with state of the art classification techniques}
\scalebox{0.75}{
\label{table:table6}
\def\arraystretch{1.5}%
\begin{tabular}{lllllll}
\hline
\hline
\multirow{2}{*}{{\bf Methods}}                     & \multicolumn{3}{c}{{\bf SVEB}}       & \multicolumn{3}{c}{{\bf VEB}}        \\
                                                   & {\bf Se} & {\bf PPV} & {\bf F-Measure} & {\bf Se} & {\bf PPV} & {\bf F-Measure} \\
\hline
Chazal et al \cite{chazal1}                               & 75.9     & 38.5      & 51.08         & 77.7     & 81.9      & 79.74         \\
Chazal et al \cite{chazal2}                                & 87.7     & 47        & 61.20         & 94.3     & 96.2      & 95.24         \\
Alvarado et al \cite{alvarado}                             & 86.19    & 56.68     & 68.38         & 92.43    & 94.82     & 93.60         \\
Ince et al \cite{ince}                                   & 63.5     & 53.7      & 58.19         & 84.6     & 87.4      & 85.97         \\
Wiens et al \cite{wiens}                                  & 92       & 99.5      & 95.60         & 99.6     & 99.3      & 99.44         \\
{\bf Proposed LDA (11 dimensional feature vector)} & 91.94    & 67.52     & 77.86         & 81.98    & 96.63     & 88.70         \\
{\bf Proposed MOE (4 dimensional feature vector)}  & 93.74    & 58.88     & 72.33         & 69.43    & 94.58     & 80.08         \\
{\bf Proposed ANN Ensemble (4 dimensional feature vector)}  & 87.19 (0.8)*    & 83.78 (1.08)*     & 85.45 (0.26)*         & 89.78 (0.18)*    & 92.56    (1.05)* & 91.14 (0.53)*  \\
      \hline
            \hline
            * Number in parenthesis is Standard Error (s.e)

\end{tabular}}
\end{table*}


\begin{thebibliography}{1}
\bibitem{WHO}
\url{http://www.who.int/cardiovascular_diseases/en/}
\bibitem{WHF}
\url{http://www.world-heart-federation.org}
\bibitem{economicimpact}
\url{http://cebp.aacrjournals.org/content/13/12/2126.full}

\bibitem{sensornetwork}
A. Volmer and R. Orglmeister, \emph{Wireless body sensor network for lowpower motion-tolerant syncronized vital sign measurment,} in Engineering in Medicine and Biology Society, 2008. EMBS 2008. 30th Annual International Conference of the IEEE, aug. 2008, pp. 3422 –3425.
\bibitem{Guyon}
Isabelle, Guyon, Andre, Elisseef, \emph{An Introduction Variable and Feature Selection}, Journal of Machine Learning Research,3, (2003), 1157-1183
\bibitem{lannoy}
Gauthier Doquire, Gael de Lannoy, Damien François, Michel Verleysen \emph{Feature Selection for Interpatient Supervised Heart Beat Classification}. Comp. Int. and Neurosc. 2011 (2011)
\bibitem{wiens}
J. Wiens and J. Guttag, \emph{Active learning applied to patient-adaptive heartbeat classification,} in Advances in Neural Information Processing Systems (NIPS), December 2011
\bibitem{ince}
T. Ince, S. Kiranyaz, and M. Gabbouj, \emph{A generic and robust system for automated patient-specific classification of ecg signals,} Biomedical Engineering, IEEE Transactions on, vol. 56, no. 5, pp. 1415--1426, (2009)
\bibitem{chazal1}
P. de Chazal, M. O’Dwyer, and R. Reilly, \emph{Automatic classification of heartbeats using ECG morphology and heartbeat interval features,} Biomedical Engineering, IEEE Transactions on, vol. 51, no. 7, pp. 1196--1206, (2004)
\bibitem{physionet}
A. L. Goldberger, L. A. N. Amaral, L. Glass, J. M. Hausdorff, P. C. Ivanov, R. G. Mark, J. E. Mietus, G. B. Moody, C.-K. Peng, and H. E. Stanley, \emph{PhysioBank, PhysioToolkit, and PhysioNet: Components of a new research resource for complex physiologic signals,} Circulation, vol. 101, no. 23, pp. e215--e220, (2000)
\bibitem{Hu}
Y. H. Hu, S. Palreddy, and W. J. Tompkins, \emph{A patient adaptable ECG beat classifier using a mixture of experts approach,} IEEE Trans. on Biomedical Engineering, vol. 44, no. 9, pp. 891--900, (1997)
\bibitem{chazal2}
P. de Chazal and R. Reilly, \emph{A patient-adapting heartbeat classifier using ECG morphology and heartbeat interval features,} Biomedical Engineering, IEEE Transactions on, vol. 53, no. 12, pp. 2535 –2543, Dec. 2006.
\bibitem{alvarado}
Alexander Singh Alvarado, Choudur Lakshminarayan, Jose C. Principe, \emph{Time-based Compression and Classification of Heartbeats}, IEEE Transactions on Biomedical Engineering, 99, (2012)
\bibitem{BDA}
Tony Basil, Bollepalli S. Chandra, and Choudur Lakshminarayan, \emph{A Comparison of Statistical Machine Learning Methods in Heartbeat Detection and Classiﬁcation,} Lecture Notes in Computer Science Volume 7678, 2012, pp 16-25
\bibitem{ACS}
Tony Basil, Choudur Lakshminarayan, and C. Krishna Mohan, \emph{Detection of Classes of Heart Arrhythmias based on Heartbeat Morphology Patterns,} SIAM 2nd International Workshop on Analytics for Cyber-Physical Systems, (2013) 
\bibitem{EUSIPCO}
Tony Basil and Choudur Lakshminarayan, \emph{Automatic Classification of Heartbeats}, European Signal Processing Conference (EUSIPCO), 2014
\bibitem{AAMI}
R. Mark and R. Wallen, \emph{AAMI-recommended practice: \emph{testing} and reporting performance results of ventricular arrhythmia detection algorithms,} Tech. Rep. AAMI ECAR, (1987)
\bibitem{elementstextbook}
Trevor Hastie, Robert Tibshirani, and Jerome Friedman. \emph{The Elements of Statistical Learning} Springer Series in Statistics Springer New York Inc., New York, NY, USA, (2001)
\bibitem{appliedtextbook}
R.A. Johnson, and D.W. Wichern. \emph{Applied Multivariate Statistical Analysis,} 3rd edition, Englewood Cliffs, New Jersey: Prentice Hall, (1992)
\bibitem{stepwise}
\url{http://en.wikipedia.org/wiki/Stepwise_regression}
\bibitem{sourceforge}
\url{https://sourceforge.net/projects/ecganalysis/}
\bibitem{chazal3}
P.de Chazal, \emph{A Switching Feature Extraction System for ECG Heartbeat Classification,} Computing in Cardiology Conference (CinC), (2013) 
\bibitem{neuralnetwork}
P.M. Granitto, P.F. Verdes, and H.A. Ceccatto,  \emph{Neural network ensembles: evaluation of aggregation algorithms,} Artificial Intelligence, (2005) 

\end{thebibliography}
\end{document}